\documentclass[preprint,12pt]{elsarticle}

\usepackage{amssymb}
\usepackage{amsmath}
\usepackage{amsthm}

\usepackage{comment}
\usepackage{algorithm}
\usepackage{algpseudocode}
\usepackage{xcolor}
\usepackage{booktabs, multirow} 
\usepackage{makecell}

\usepackage{hyperref}
\hypersetup{breaklinks=true}
\hypersetup{nolinks=true}

\journal{Information Fusion}

\begin{document}

\newcommand{\TODO}[1]{\textbf{{\color{red} !!!! TODO: #1 !!!! }}}
\newcommand{\TOREF}{\TODO{ADD REFERENCE}}

\begin{frontmatter}



\title{Do Multimodal Large Language Models Understand Welding?}


\author[a]{Grigorii Khvatskii} 
\author[b]{Yong Suk Lee\corref{cor1}}
\author[c]{Corey Angst}
\author[d]{Maria Gibbs}
\author[e]{Robert Landers}
\author[a]{Nitesh V. Chawla\corref{cor1}}

\cortext[cor1]{Corresponding authors. Email: yong.s.lee@nd.edu; nchawla@nd.edu}

\affiliation[a]{organization={Department of Computer Science and Engineering, University of Notre Dame},
            city={Notre Dame},
            postcode={46556}, 
            state={IN},
            country={USA}}
\affiliation[b]{organization={Keough School of Global Affairs, University of Notre Dame},
            city={Notre Dame},
            postcode={46556}, 
            state={IN},
            country={USA}}
\affiliation[c]{organization={Information Technology, Analytics, and Operations Department, University of Notre Dame},
            city={Notre Dame},
            postcode={46556}, 
            state={IN},
            country={USA}}
\affiliation[d]{organization={Industry Labs, University of Notre Dame},
            city={Notre Dame},
            postcode={46556}, 
            state={IN},
            country={USA}}
\affiliation[e]{organization={Department of Aerospace and Mechanical Engineering, University of Notre Dame},
            city={Notre Dame},
            postcode={46556}, 
            state={IN},
            country={USA}}
            
\begin{abstract}

This paper examines the performance of Multimodal LLMs (MLLMs) in skilled production work, with a focus on welding. Using a novel data set of real-world and online weld images, annotated by a domain expert, we evaluate the performance of two state-of-the-art MLLMs in assessing weld acceptability across three contexts: RV \& Marine, Aeronautical, and Farming. 
While both models perform better on online images, likely due to prior exposure or memorization, they also perform relatively well on unseen, real-world weld images. 
Additionally, we introduce WeldPrompt, a prompting strategy that combines Chain-of-Thought generation with in-context learning to mitigate hallucinations and improve reasoning. 
WeldPrompt improves model recall in certain contexts but exhibits inconsistent performance across others. These results underscore the limitations and potentials of MLLMs in high-stakes technical domains and highlight the importance of fine-tuning, domain-specific data, and more sophisticated prompting strategies to improve model reliability. The study opens avenues for further research into multimodal learning in industry applications.

\end{abstract}


\begin{highlights}
\item We evaluate MLLMs' performance in assessing weld quality.
\item We introduce WeldPrompt, a strategy using Chain-of-Thought and in-context learning.
\item MLLMs perform better on online than real-world weld images, showing limited generalization.
\item WeldPrompt boosts recall in some contexts but trades off precision in different applications.
\item MLLM limitations in welding offer insights for future XAI research in manufacturing.
\end{highlights}


\end{frontmatter}



\section{Introduction}

With the advent of generative artificial intelligence (Gen AI), inaccurate or false model output, often referred to as hallucinations, has become an expanding area of research \cite{nahar2024}. While the topic continues to be widely researched \cite{10.1145/3571730, rawte-etal-2023-troubling, 10.1145/3583780.3614905}, existing research on large language model (LLM) hallucinations has mostly focused on object hallucinations, particularly in text \cite{bai2024}, leaving some emerging questions understudied. One such question is the prevalence of errors and hallucinations in multimodal LLMs, especially when presented with previously unseen real-world images and in novel domains. Hallucinations in multimodal LLMs may be a greater concern than those in text-only LLMs due to the increased number of potential failure points that lead to false or incorrect outputs \cite{wang2023}. In addition, these LLMs can be used in higher-stakes settings, such as self-driving cars or healthcare data analysis, making this problem an urgent area of study.

While an increasing number of papers have examined the impact of LLMs on the labor market \cite{doi:10.1126/science.adj0998}, they have disproportionately focused on the potential impact on white-collar or office workers. However, the effects of the wide adoption of LLMs on skilled production workers have received considerably less attention from researchers. Since skilled production work often deals with physical objects, this segment of the workforce presents a unique opportunity to integrate multimodal LLMs that can work with images in their workflows \cite{Makatura2024Large}.

In this work, we assess the performance of multimodal LLMs in the context of skilled production work, with a specific focus on welding. For this, we have collected a unique dataset of weld images, combining images available online with a new set of real-world welding images collected from shop floors and training centers. These images were then annotated by a welding expert, who assessed whether the welds were acceptable for a diverse set of applications. Additionally, we developed a novel LLM prompting strategy (WeldPrompt) that combines automatic Chain-of-Thought generation with in-context learning to mitigate LLM hallucinations. Using few-shot in-context learning was shown in prior work to reduce the prevalence of model hallucinations. 

In our study, we assess how closely the model's outputs align with expert judgments. This approach diverges from conventional definitions of LLM hallucinations (models outputting fictitious information \cite{selfcheckgpt}) due to the subjective nature of our context and the challenge of establishing a singular ground truth. Conventional hallucination detection techniques are currently lacking \cite{malin2024reviewfaithfulnessmetricshallucination,ravichander2025halogenfantasticllmhallucinations} in such subjective open-ended tasks. We examine the performance of LLMs in identifying acceptable welds using a diverse set of metrics that measure classification performance, which, in our case, is a proxy for expert alignment.

The rest of the paper is organized as follows. In Section \ref{sec:rw} we provide a short overview of the existing applications of Large Language Models in manufacturing. In Section \ref{sec:mm}, we describe our data collection procedure, as well as the pipeline used to generate LLM responses and evaluate model performance. In Section \ref{sec:tc}, we present the results of the experimental evaluation. Finally, we discuss the implications and conclude in the last section.

\section{Related work}
\label{sec:rw}

While the effects of generative AI on the modern society and economy have been widely discussed in the literature \cite{hui_short-term_2024, lin_hiding_2024, woodruff_how_2024, layman_generative_2024, john_villasenor_generative_2024}, current research typically focuses on knowledge-driven (e.g. white-collar work, education) and creative (e.g. art) contexts. At the same time, the discussion of Gen AI use in more physical contexts (for example, manufacturing) remains much smaller, even though existing literature hints at its large potential \cite{shi_leveraging_2024}. 

Existing research focuses heavily on text-based LLMs, with a particular focus on adding context-relevant knowledge to the models. For example, Retrieval-Augmented Generation was utilized in combination with a pretrained LLM to aid in answering questions related to Additive Manufacturing \cite{chandrasekhar_amgpt:_2024}. A similar text-based pipeline has shown promise in accelerating material science discoveries \cite{liu_prompt-engineered_2024}. 

The explored uses of LLMs in manufacturing go beyond strict research and development use. Fine-tuned LLMs have shown proficiency in generating manufacturing domain-specific code \cite{xia_leveraging_2024}, as well as answering other domain-specific questions. Such systems with access to domain knowledge have also shown promise for quality control in the aerospace industry \cite{zhou_causalkgpt:_2024}.

All of these studies, however close to the workshop floor, have only utilized text and graph data in their pipelines. However, computer vision has a large history of being used in manufacturing \cite{newman_survey_1995, rao_future_1996, zhou_computer_2023, qamar_application_2024}, as well as other industries, such as agriculture \cite{molto_computer_1999, tian_computer_2020}, where it was widely used to allow interactions between software and physical objects. This suggests that combining computer vision and large language models may benefit existing manufacturing processes, as well as allowing novel ones. 

While a significant amount of work was done on welding assesment using more well-established approaches to computer vision \cite{amarnath_automatic_2023}, the use of generative multimodal models in this context remains understudied. Deep learning based approaches were used manufacturing \cite{kothari_detecting_2018}, construction \cite{shafeek_assessment_2004}, and welding education \cite{ngoc_enhancing_2024}, where they all have shown promising results. An important feature of using MLLMs in this context is their ability to provide more detailed feedback on the welds, including explanations of the reasioning behind their decisions.

Practical implementations of such pipelines became feasible after multimodal large language models (MLLMs) were introduced. MLLMs allow for both text and image (or video input) \cite{bai_survey_2024}, thus allowing to build pipelines effectively combining computer vision and text generation. These models have been succesfully applied in some contexts. For example, they have been used for medical decision support \cite{liu_medical_2023}. At the same time, this combination remains understudied in the context of manufacturing, where it might be immediately applicable.

\section{Material and methods}
\label{sec:mm}

In this section, we outline the data collection procedure used to build the datasets for our experiments. We then describe the methodology used to perform the experimental evaluation of LLM performance.

\subsection{Datasets}


For our experiments, we collected and labeled two data sets of weld images. The first dataset, the Real-World Weld Dataset, was gathered directly from expert welders with experience in manufacturing production environments who serve as instructors in welding apprenticeship and training programs. Expert welders were instructed to take photos of welds representing easy, moderate, and difficult welding problems and then provide a narrative describing the weld and what would need to be done to correct the problem. The second dataset, the Online Weld Dataset, was assembled by downloading publicly available weld images of welds from the internet. In assembling the Online Weld Dataset, a systematic approach was adopted to collect a diverse range of weld images. Data collection involved conducting targeted Google image searches using specific keywords such as "welding problems," "weld defects," "bad welds", "good welds," and "welding image example." When freely available, images from welding manuals and guides were prioritized to ensure high-quality and relevant content. A careful selection process was implemented to ensure that the images were high-quality photographs of actual welds, excluding any graphic or drawn representations and avoiding images with text overlays. This methodology facilitated the acquisition of a set of 73 images representing various welding conditions and techniques. These two datasets were collected to evaluate the LLM's performance on data it might have encountered during its training process, as well as completely new, unseen data.

After collection, both data sets were labeled by a domain expert, who were asked to evaluate the acceptability of the pictured welds in different contexts. For this work, the expert has evaluated the acceptability of the welds in the contexts of RV \& Marine, Aeronautical, and Farming. These contexts were chosen because of their diverse requirements for welding techniques and the varying tolerances to common mistakes. For each welding context, the expert has provided a binary response indicating whether this weld is acceptable, as well as a detailed description of why the weld was deemed acceptable or not. The descriptive statistics of the labeled data set are provided in Table \ref{tab:dsinfo}. We can see that in both the Real-World and Online datasets, the Aeronautical context displays large data imbalance, due to the more stringent standards required in aerospace vehicles.


\begin{table}[!htp]\centering
\scriptsize
\begin{tabular}{lrrrrrrr}\toprule
\multirow{2}{*}{\textbf{}} &\multicolumn{3}{c}{\textbf{Real World}} &\multicolumn{3}{c}{\textbf{Online}} \\\cmidrule{2-7}
&\textbf{POS} &\textbf{NEG} &\textbf{IMB} &\textbf{POS} &\textbf{NEG} &\textbf{IMB} \\\midrule
RV / Marine &43 &52 &0.827 &35 &23 &1.522 \\
Aeronautical &12 &83 &0.145 &16 &42 &0.381 \\
Farming &26 &69 &0.377 &41 &17 &2.412 \\
\bottomrule
\end{tabular}
\caption{Descriptive statistics of the datasets}
\label{tab:dsinfo}
    \vspace{1em} 
    \parbox{0.8\textwidth}{\footnotesize Note: POS stands for images that were labeled acceptable by the domain expert, NEG stands for images labeled unacceptable, and IMB stands for imbalance ratio, which we define as the number of positive instances divided by negative instances.}
\end{table}

Some of the images in both datasets contained various annotations (such as "CRACK"), either as something physically written on the samples of welds themselves, or added digitally on top of original images. Such annotations have been shown to affect MLLM output \cite{amara2024contextmattersvqareasoning}, through MLLMs relying on them for classification. To avoid this, we visually inspected all the images in our datasets and removed images containing such annotations. This has allowed us to force the models to focus on classifyng the actual welds instead of any extraneous annotations. In total, we have removed 15 images from the Online dataset and 11 images from the Real world dataset.

\subsection{Experimental Setting}

In our experiments, we tested the ability of GPT-4o and LLaVA-1.6 MLLMs to evaluate the acceptability of the weld in different contexts. We evaluated the model's performance in two settings: zero-shot and chain-of-thought prompting, which we refer to as WeldPrompt.

In the zero-shot setting, the model received only the weld image as input and was prompted to generate a binary response about the weld's acceptability in various contexts, along with a short explanation for its reasoning.

When using WeldPrompt, the model was provided with access to weld images and descriptions of similar welds that had led to correct zero-shot answers. The model was then asked to provide short explanations and binary responses regarding the acceptability of the weld in various contexts.

While these methods may underutilize model context by not filling it completely, recent studies \cite{an2024doeseffectivecontextlength} show that the relationship between context utilization and model performance is nonlinear, even for long-context models. Additionally, while GPT-4o was trained to utilize a 128000 tokens context depth, LLaVA-1.6 has an effective context size of only 32768 tokens. Finally, reliance on large contexts may result in models consuming a large amount of memory, making methods that don't need large contexts a lucrative target for edge device deployment. Furhermore, our approach is closer to how a welder might realistically interact with large language models in a real-world setting, especially when considering the use of locally deployed MLLMs.

\subsubsection{Zero-Shot prompting}

In our zero-shot model evaluation experiments, we prompted the model to generate image evaluations similar to those provided by the expert. For each image, we first queried the model to describe characteristics of the weld that could affect its acceptability in the given context. We then 
asked the model to generate a binary answer indicating whether the weld is acceptable in the given context or not.

\begin{algorithm}
\caption{Zero-shot model prompting algorithm}
\begin{algorithmic}
\Require Weld Image
\For{r in R \text{runs}}
\For {c in C \text{contexts}}
\State $M_{r,c} \gets \{\text{System Message}(c)\}$
\State $M_{r,c} \gets M_{r,c} + \{\text{Reasoning prompt}(c), \text{Weld Image}\}$
\State $R_{r,c} \gets \textbf{MLLM}(M_{r,c})$ \Comment{Generate model reasoning}
\State $M_{r,c} \gets M_{r,c} + \{R_{r,c}, \text{Binary Answer Propmt}(c)\}$
\State $B_{r,c} \gets \textbf{MLLM}(M_{r,c})$ \Comment{Generate binary response}
\EndFor
\EndFor
\State \Return $[\{R_{r,c}, B_{r,c}\}$ for all $r, c]$
\end{algorithmic}
\label{alg:zshp}
\end{algorithm}

In our evaluation procedure, each context was evaluated independently, with no information transferred between the weld acceptability evaluations across different contexts. For our experiments, we conducted three separate model prompting runs for each image, using different random seeds. A detailed description of the prompting steps is provided in Algorithm.\ref{alg:zshp}.

\subsubsection{WeldPrompt}

In contrast to zero-shot prompting, where the model generates image evaluations without prior information, we employed a procedure based on MedPrompt \cite{medprompt}, allowing the model to access information about similar welds in the dataset. 

First, we generated the model's reasoning responses and binary answers using Algorithm \ref{alg:zshp}. For each image, we computed its embedding using the \texttt{vit-base-patch16-224} pre-trained model. For each image and context, we then identified the runs where the model's answer was correct, saving the corresponding reasoning responses, correct binary answers, and image embeddings. This pre-processing procedure is described in Algorithm \ref{alg:mppp}.

\begin{algorithm}
\caption{Chain-of-thought image data preprocessing}
\begin{algorithmic}
\Require {Reference images}
\For {i in \text{Reference images}}
\State $E_i \gets \textbf{vit-base-patch16-224}(i)$
\State $[R_{i,c}, B_{i,c}] \gets \text{Algorithm \ref{alg:zshp}}(i)$
\For {c in C \text{contexts}}
\State $C_{i,c} \gets \{\}$ 
\For{r in R \text{runs}}
\If {$B_{i,r,c} \text{ is correct}$}
\State $C_{i,c} \gets C_{i,c} + \{(R_{i,r,c}, {B_i,r,c})\}$
\EndIf
\EndFor
\EndFor
\EndFor
\State \Return $[\{i, E_{i}, C_{i,c}\}$ for all $i, c]$
\end{algorithmic}
\label{alg:mppp}
\end{algorithm}

To evaluate the model's performance, we computed an embedding for each input image using the \texttt{vit-base-patch16-224} pre-trained model. We then used cosine similarity to identify the five closest images in the pre-processed reference set. For our experiments, we used a leave-one-out approach, where the reference set for each input image included the entire dataset except for the input image itself.

Next, for each image, we added the reasoning responses, correct binary answers, and images retrieved in the previous step into the model context. By doing so, we supplied the model with additional task examples and chains-of-thought, allowing us to leverage its in-context learning capabilities and enhance its classification performance. We then employed a prompting procedure similar to the one described in the previous section, querying the model for reasoning and a binary response. This procedure was performed in three runs for each image context. The pseudocode description of the prompting strategy is provided in Algorithm \ref{alg:mpii}.


\begin{algorithm}
\caption{Chain-of-thought model prompting algorithm}
\begin{algorithmic}
\Require {Weld Image, Reference images}
\State $[i, E_{i}, C_{i,c}] \gets \text{Algorithm \ref{alg:mppp}}
(\text{Reference Images})$
\State $F \gets \textbf{vit-base-patch16-224}(\text{Weld Image})$
\State $[\hat{i}, \hat{E_{i}}, \hat{C_{i,c}}] \gets \text{Find 5 Closest}(F, E_i)$ \Comment{Identify 5 closest images}
\For{r in R \text{runs}}
\For {c in C \text{contexts}}
\State $M_{r,c} \gets \{\text{System Message}(c)\}$
\State $M_{r,c} \gets M_{r,c} + \{\text{Chain-Of-Thought Prompt}(\hat{C_{i,c}} , \hat{i})\}$
\State $K_{r,c} \gets \textbf{MLLM}(M_{r,c})$ \Comment{Add Chain-Of-Thought}
\State $M_{r,c} \gets M_{r,c} + \{K_{r,c}, \text{Reasoning Prompt}(c)\}$
\State $R_{r,c} \gets \textbf{MLLM}(M_{r,c})$ \Comment{Generate model reasoning}
\State $M_{r,c} \gets M_{r,c} + \{R_{r,c}, \text{Binary Answer Prompt}(c)\}$
\State $B_{r,c} \gets \textbf{MLLM}(M_{r,c})$ \Comment{Generate binary response}
\EndFor
\EndFor
\State \Return $[\{R_{r,c}, B_{r,c}\}$ for all $r, c]$
\end{algorithmic}
\label{alg:mpii}
\end{algorithm}

\subsubsection{Evaluation procedure}

After generating the reasoning and binary responses for both zero-shot and WeldPrompt setting, we computed several evaluation metrics for model performance on both the Real-World and Online datasets.
First, we used the binary answers and expert evaluations to compute multilabel classification performance metrics. For each image, we used majority vote across all runs to determine the binary response for evaluation. For ROC-AUC (Area Under Receiver Operating Characteristic Curve) computation, we averaged the binary responses across all runs to estimate probabilities. We computed precision, recall, F1-score, and ROC-AUC for each weld context separately, and also computed the micro, macro, and sample averages of these scores.




\section{Results}
\label{sec:tc}

\subsection{Experimental Results}
\label{ssec:er}

\subsubsection{Classification performance}
\label{sssec:class}

We evaluated the zero-shot classification performance of the \texttt{GPT-4o} and \texttt{LLaVA-1.6} \footnote{We have used the \texttt{LLaVa-1.6-Mistral-7B} model at \texttt{Q8\_0} quantization} Multimodal Large Language Models (MLLMs) in RV \& Marine, Aeronautical, and Farming contexts by computing precision, recall, F1, and ROC-AUC measures from the model's binary responses. We assessed the model's performance on both the Real-World and Web datasets. The evaluation results are shown in Table \ref{tab:0S_CLASS_REPORT_VARS}.

\begin{table}[!htp]\centering
\scriptsize
\begin{tabular}{lrrrrrrr}
\toprule
& &\multicolumn{2}{c}{\textbf{RV \& Marine}} &\multicolumn{2}{c}{\textbf{Aeronautical}} &\multicolumn{2}{c}{\textbf{Farming}} \\\cmidrule{3-8}
\textbf{} & &\makecell{\textbf{Real} \\ \textbf{world}} &\textbf{Online} &\makecell{\textbf{Real} \\ \textbf{world}} &\textbf{Online} &\makecell{\textbf{Real} \\ \textbf{world}} &\textbf{Online} \\\midrule
\multirow{2}{*}{\textbf{PREC}} &GPT-4o &0.688 &0.846 &0.625 &0.636 &0.383 &0.857 \\
&LLaVA-1.6 &0.455 &0.778 &0.250 &0.556 &0.250 &0.929 \\
\multirow{2}{*}{\textbf{REC}} &GPT-4o &0.512 &0.629 &0.417 &0.438 &0.692 &0.585 \\
&LLaVA-1.6 &0.116 &0.400 &0.167 &0.313 &0.231 &0.317 \\
\multirow{2}{*}{\textbf{F1}} &GPT-4o &0.587 &0.721 &0.500 &0.519 &0.493 &0.696 \\
&LLaVA-1.6 &0.185 &0.528 &0.200 &0.400 &0.240 &0.473 \\
\multirow{2}{*}{\textbf{ROCAUC}} &GPT-4o &0.710 &0.745 &0.727 &0.779 &0.644 &0.693 \\
&LLaVA-1.6 &0.522 &0.560 &0.604 &0.688 &0.462 &0.681 \\
\bottomrule
\end{tabular}
\caption{Zero-Shot Classification Performance per variable}
\label{tab:0S_CLASS_REPORT_VARS}
\end{table}

From the table, we can see that the models show similar between different context. Both models generally underperformed in the Farming context. 

The relationship between precision and recall in the Aeronautical context suggests that the GPT-4o model was generally conservative, rejecting welds deemed acceptable by the expert. A similar, though less pronounced, relationship between precision and recall exists for the RV \& Marine context. For the Farming context, however, the relationship is reversed, indicating that the model was more lenient than the expert when labeling the welds. LLaVA-1.6 displayed similar properties, also having its recall higher that precision. 
The trade-off between precision and recall was greater in models for the Aeronautical context as well. 

GPT-4o was stricter in its predictions, often favoring precision over recall, which led to fewer false positives. LLaVA-1.6 has generally underperformed in our experiments, which is to be excpected, given its much smaller size.

In terms of overall classification performance, both models performed better on the Online dataset compared to the Real-World dataset, with GPT-4o outperforming LLaVa-1.6 in all contexts. This is evident from looking both at the performance in individual contexts, as well as from comparing performance averages show in Table \ref{tab:0S_CLASS_REPORT_AVGS}. This is to be expected and suggests that the model relies heavily on memorizing and retrieving its training data obtained from online sources, instead of using reasoning on the given images.

\begin{table}[!htp]\centering
\scriptsize
\begin{tabular}{lrrrrrrrrrr}\toprule
& &\multicolumn{2}{c}{\textbf{MIC}} &\multicolumn{2}{c}{\textbf{MAC}} &\multicolumn{2}{c}{\textbf{WEIG}} &\multicolumn{2}{c}{\textbf{SAMP}} \\\cmidrule{3-10}
\textbf{} & &\makecell{\textbf{Real} \\ \textbf{world}} &\textbf{Online} &\makecell{\textbf{Real} \\ \textbf{world}} &\textbf{Online} &\makecell{\textbf{Real} \\ \textbf{world}} &\textbf{Web} &\makecell{\textbf{Real} \\ \textbf{world}} &\textbf{Online} \\\midrule
\multirow{2}{*}{\textbf{PREC}} &GPT-4o &0.517 &0.815 &0.565 &0.780 &0.580 &0.815 &0.247 &0.402 \\
&LLaVA-1.6 &0.302 &0.780 &0.318 &0.754 &0.359 &0.806 &0.104 &0.362 \\
\multirow{2}{*}{\textbf{REC}} &GPT-4o &0.556 &0.576 &0.540 &0.550 &0.556 &0.576 &0.277 &0.351 \\
&LLaVA-1.6 &0.160 &0.348 &0.171 &0.343 &0.160 &0.348 &0.082 &0.210 \\
\multirow{2}{*}{\textbf{F1}} &GPT-4o &0.536 &0.675 &0.527 &0.645 &0.544 &0.675 &0.249 &0.368 \\
&LLaVA-1.6 &0.210 &0.481 &0.208 &0.467 &0.205 &0.481 &0.087 &0.256 \\
\bottomrule
\end{tabular}
\caption{Zero-Shot Classification Performance averages}
\label{tab:0S_CLASS_REPORT_AVGS}
\end{table}

We also evaluated the classification performance of \texttt{GPT-4o} and LLaVA-1.6 using WeldPrompt in the RV \& Marine, Aeronautical, and Farming contexts by computing the same set of classification performance measures from the model's binary responses. As with the zero-shot setting, we assessed the model's performance on both the Real-World and Online datasets. The results of the evaluation are shown in Table \ref{tab:MP_CLASS_REPORT_VARS}.

In the RV \& Marine context, the introduction of WeldPrompt led to an improvement in both GPT-4o’s recall and precision, indicating an overall increase in classification performance. At the same time, in the RV \& Marine context, GPT-4o’s precision increased at the cost of recall. In general, LLaVa-1.6 has shown a similar dynamic when comparing 0-shot results with WeldPrompt.

The results show a dynamic similar to the zero-shot setting, where the GPT-4o model was considerably stricter than the expert in the Aeronautical context, that is, it was rejecting samples deemed acceptable by the expert. However, with WeldPrompt, this strictness also extended to the Farming context.


For the Real-World images in the Aeronautical context, LLaVA-1.6 showed zero classification performance on the Real-World dataset, likely due to the large class imbalance, resulting in the model classifying all images as unacceptable. 
However, the GPT-4o model is less affected by the class imbalance and more aggressively predicts welds to be acceptable compared to the LLaVA-1.6 model in all three contexts. 

Focusing on the Real-World data set, when we compare WeldPrompt to the zero-shot setting, we generally see improvements in both models's F1 scores, suggesting that WeldPrompt has made better aligned with the domain expert. While the performance of the models decreased in the Aeronautical context, this was compensated by significant performance increases in the other contexts. The Farming context displays an interesting dynamic, where the use of WeldPrompt increased precision and decreased recall, although resulting in an overall increase in F1 performance.

In terms of average performance (shown in Table \ref{tab:MP_CLASS_REPORT_AVGS}), applying WeldPrompt to both models led to a slight decrease in performance on the Real-World dataset but increased classification performance on the Online dataset. This reinforces our conclusion that the model relies heavily on memorization rather than reasoning when classifying welding images.

\begin{table}[!htp]\centering
\scriptsize
\begin{tabular}{lrrrrrrrr}\toprule
& &\multicolumn{2}{c}{\textbf{RV \& Marine}} &\multicolumn{2}{c}{\textbf{Aeronautical}} &\multicolumn{2}{c}{\textbf{Farming}} \\\cmidrule{3-8}
\textbf{} & &\makecell{\textbf{Real} \\ \textbf{world}} &\textbf{Online} &\makecell{\textbf{Real} \\ \textbf{world}} &\textbf{Online} &\makecell{\textbf{Real} \\ \textbf{world}} &\textbf{Web} \\\midrule
\multirow{2}{*}{\textbf{PREC}} &GPT-4o &0.700 &0.885 &0.286 &0.750 &0.390 &0.871 \\
&LLaVA-1.6 &0.667 &0.778 &0.000 &0.333 &0.286 &0.885 \\
\multirow{2}{*}{\textbf{REC}} &GPT-4o &0.651 &0.657 &0.167 &0.750 &0.615 &0.659 \\
&LLaVA-1.6 &0.186 &0.600 &0.000 &0.188 &0.077 &0.561 \\
\multirow{2}{*}{\textbf{F1}} &GPT-4o &0.675 &0.754 &0.211 &0.750 &0.478 &0.750 \\
&LLaVA-1.6 &0.291 &0.677 &0.000 &0.240 &0.121 &0.687 \\
\multirow{2}{*}{\textbf{ROCAUC}} &GPT-4o &0.727 &0.772 &0.620 &0.824 &0.677 &0.741 \\
&LLaVA-1.6 &0.575 &0.698 &0.412 &0.569 &0.481 &0.722 \\
\bottomrule
\end{tabular}
\caption{WeldPrompt Classification Performance per variable}
\label{tab:MP_CLASS_REPORT_VARS}
\end{table}

\begin{table}[!htp]\centering
\scriptsize
\begin{tabular}{lrrrrrrrrrr}\toprule
& &\multicolumn{2}{c}{\textbf{MIC}} &\multicolumn{2}{c}{\textbf{MAC}} &\multicolumn{2}{c}{\textbf{WEIG}} &\multicolumn{2}{c}{\textbf{SAMP}} \\\cmidrule{3-10}
\textbf{} & &\makecell{\textbf{Real} \\ \textbf{world}} &\textbf{Online} &\makecell{\textbf{Real} \\ \textbf{world}} &\textbf{Online} &\makecell{\textbf{Real} \\ \textbf{world}} &\textbf{Web} &\makecell{\textbf{Real} \\ \textbf{world}} &\textbf{Online} \\\midrule
\multirow{2}{*}{\textbf{PREC}} &GPT-4o &0.523 &0.849 &0.459 &0.835 &0.539 &0.855 &0.309 &0.443 \\
&LLaVA-1.6 &0.385 &0.758 &0.317 &0.665 &0.446 &0.748 &0.086 &0.425 \\
\multirow{2}{*}{\textbf{REC}} &GPT-4o &0.568 &0.674 &0.478 &0.689 &0.568 &0.674 &0.314 &0.414 \\
&LLaVA-1.6 &0.123 &0.511 &0.088 &0.449 &0.123 &0.511 &0.088 &0.330 \\
\multirow{2}{*}{\textbf{F1}} &GPT-4o &0.544 &0.752 &0.454 &0.751 &0.543 &0.752 &0.297 &0.418 \\
&LLaVA-1.6 &0.187 &0.610 &0.137 &0.535 &0.193 &0.605 &0.084 &0.355 \\
\bottomrule
\end{tabular}
\caption{WeldPrompt Classification Performance averages}
\label{tab:MP_CLASS_REPORT_AVGS}
\end{table}

\section{Discussion}
The results of this study provide several important insights into the performance of multimodal Large Language Models (MLLMs), particularly in the context of skilled production work, like welding. We discuss the key takeaways from our analysis and their implications.\\
\\
\noindent \textit{Performance Differences between Real-World vs. Online Images}: The models generally perform better on online images than on unseen, real-world images. While this suggests that MLLMs like GPT-4o and LLaVA-1.6 may rely on memorization of training data, it’s notable that they still perform reasonably well on real-world images they’ve never encountered. This reflects a growing capability for handling unfamiliar data, which we may sometimes overlook given the models’ successes in more familiar tasks. The better performance on online images likely stems from the resemblance to those seen during training, but the weaker performance on real-world weld images highlights a gap in welding-specific reasoning and the ability to generalize effectively to novel contexts. This underscores the need for further fine-tuning and more robust reasoning strategies in industrial applications.\\
\\
\noindent \textit{Multimodal Performance Limitations}: LLMs have shown incredible success in making text-based predictions and statements on domains that it was not particularly trained for. Relative to the success observed in text, MLLMs seem to struggle more in understanding and making predictions based on images from domains such as welding. The finding that these models can still perform reasonably well on unseen, real-world images is noteworthy, and points to emerging capabilities in understanding physical objects and unfamiliar domains. While current MLLMs have limitations in technical, image-based reasoning, we see potential for improvement in handling real-world data in technical domains that involve physical objects.\\
\\

\noindent \textit{Fine-Tuned Models vs. General-Purpose MLLMs}:  Smaller models fine-tuned for specific image-related tasks have historically shown potential to outperform more general-purpose MLLMs such as GPT-4o in specific contexts. However, recent experiments indicate that LLaVA-1.6 has underperformed across the board when compared to GPT-4o. This suggests that further testing is required when preparing small models for production deployments in industrial applications. \\
\\

\noindent \textit{WeldPrompt vs. Zero-Shot}: The WeldPrompt method performed better than zero-shot prompting in some contexts but not in others, and depended on the model used. 
When comparing WeldPrompt to the zero-shot setting on the Real-World dataset, GPT-4o showed improved F1 scores due to higher recall, suggesting WeldPrompt made the model aligned with the domain expert. However, precision decreased in the RV \& Marine and Aeronautical contexts, while recall increased. In contrast, in the Farming context, WeldPrompt improved precision but reduced recall, leading to an overall drop in F1 performance. For LLaVa-1.6, WeldPrompt slightly reduced F1 scores when compared to the zero-shot setting.\\
\\
\noindent \textit{Comparison to Other Domains}: When compared to studies in other domains, such as question answering, where pipelines like MedPrompt have shown success, our results are not as impressive. MedPrompt’s performance in question answering may be due to visual patterns between in weld images. In welding the visual patterns and physical environment as well as the use contexts and evaluation criteria are more diverse. Addressing these challenges may require domain-specific data augmentation strategies or hybrid models that combine visual reasoning with expert knowledge.
\\

\noindent \textit{Comparison to Other Computer Vision Methods}: Compared to state-of-the-art methods in the literature, such as those presented in \cite{ngoc_enhancing_2024}, which have shown an F1 score of 0.8 for defect detection, MLLM-based methods underperform in general. Given their potential, this indicates a clear need for further development and refinement. It is important to note that in our testing, we used images taken under a multitude of non-ideal conditions (varying lighting, and different angles, varying positions, sizes and shapes of samples), which might have contributed to the low model performance. This indicates the need to focus on enhancing MLLM robustness to real-world conditions.

\section{Conclusion}
Our findings highlight the limitations and potentials of current multimodal large language models (MLLMs) in industrial domains, such as welding, where precision and safety are critical. 
While the models perform better on familiar online images, the finding that they perform decently on unseen real-world weld images is notable. This capability illustrates the significant advancement of LLMs and holds promise for further improvement in handling unfamiliar data.
The performance gaps we observed in real-world vs online images in welding illustrate some current constraints of multimodal models in its applicability to real-world, high-stakes industrial contexts.
At the same time, the current limitations can inform future research directions on how to better develop explainable AI (XAI) in the context of Industry 4.0 and 5.0 and integrate explainable AI (XAI) in the context of Industry 4.0 and 5.0. 
Our findings also suggest that context-specific prompt engineering, such as WeldPrompt, may enhance performance in certain cases, though trade-offs between precision and recall remain a key consideration. 

We also analyzed the reasoning behind the model's assessments of weld acceptability compared to expert evaluations. While this analysis did not yield significant findings, further investigation into this aspect will be vital for advancing XAI research. This is especially relevant as the industry continues to develop models that not only perform better but also offer clear explanations of their decisions.

The mixed success of WeldPrompt and the models' lower performance on the real-world data suggest that future research should focus on improving MLLMs' ability to reason in unfamiliar domains, potentially through retrieval-augmented methods, in-context learning, or domain-specific fine-tuning. This aligns with the broader goals of XAI in Industry 4.0 and 5.0, which emphasize making AI systems not only more capable but also more interpretable and aligned with human decision-making.

Our results also show that model size or complexity does not necessarily lead to better performance across all contexts. Fine-tuned models tailored for specific industrial tasks may offer more reliable and cost-effective solutions compared to large, general-purpose MLLMs. As AI adoption in industrial contexts grows, practitioners will need to balance the trade-offs between using complex, expensive models and opting for task-specific models that better meet industry needs. The integration of XAI into these decision-making frameworks will be critical to ensuring that AI systems in Industry 5.0 are not only effective but also transparent and accountable.

\section*{Contributions}
\textbf{Grigorii Kvatskii}: Data curation, Formal analysis, Investigation, Methodology, Software, Writing - original draft.
\textbf{Yong Suk Lee}: Conceptualization, Data curation, Funding acquisition, Investigation, Methodology, Project administration, Resources, Supervision, Writing - original draft, Writing - review and editing.
\textbf{Corey Angst}: Conceptualization, Funding acquisition, Writing - review and editing.
\textbf{Maria Gibbs}: Data curation, Investigation, Project administration, Resources, Writing - review and editing.
\textbf{Robert Landers}: Conceptualization, Funding acquisition, Writing - review and editing.
\textbf{Nitesh Chawla}: Conceptualization, Funding acquisition, Methodology, Resources, Supervision, Writing - review and editing.

\section*{Declaration of competing interest}
The authors declare that they have no known competing financial interests or personal relationships that could have appeared to influence the work reported in this paper.

\section*{Data availability}
Data and code used in our experimental evaluation are available at the following link: \url{https://anonymous.4open.science/r/FutureOfWork-LLM-2557}

\section*{Acknowledgments}
This paper is based supported in part by the National Science Foundation under Grant Number 2222751.
The authors thank Emma Jackson, Nicoleta Paladi, Mikheil Parunovi for excellent research assistance throughout the project. 
We thank John Krause of Elkhart Area Career Center, John Rowe of Plymouth High School, Doug Bernhard of Career Academy South Bend, Jeremy Lucas of Plumbers \& Pipefitters Local Union 172, Cole Keller of Ivy Tech Community College and Satyapragnya Kar for sharing their expertise in classifying and contributing the weld images used in this study. 
Any opinions, findings, conclusions or recommendations expressed in this material are those of the authors and do not necessarily reflect the views of the National Science Foundation.

\bibliographystyle{elsarticle-num}
\bibliography{references}




\end{document}